\documentclass[runningheads]{llncs}
\usepackage[T1]{fontenc}
\usepackage{amsmath,amsfonts,bm}
\usepackage{graphicx}
\usepackage{subfig}
\usepackage{color}

\newcommand{\R}{\mathbb{R}}
\newcommand{\dom}{\mathrm{D}}
\newcommand{\lra}{\longrightarrow}

\newcommand{\regu}{\mathrm{reg}}

\begin{document}

\title{Enhancing CNNs robustness to occlusions with
bioinspired filters for border completion}
\titlerunning{Enhancing CNNs}

\author{Catarina P. Coutinho\inst{1}\orcidID{0000-0003-3707-8795} \and
Aneeqa Merhab\inst{2}\orcidID{0009-0004-8013-3061} \and
Janko Petkovic\inst{3}\orcidID{0009-0009-7455-9484} \and
Ferdinando Zanchetta \inst{1}\orcidID{0000-0003-2294-2755}  \and
Rita Fioresi \inst{1}\orcidID{0000-0003-4075-7641}
}
\authorrunning{Coutinho et al.}
%
\institute{FaBiT, University of Bologna, Via San Donato 15, 40127 Bologna, Italy
\email{catarina.praefke2@unibo.it,}
\email{rita.fioresi@unibo.it,}
\email{ferdinando.zanchett2@unibo.it} \and
Department of Mathematics, University of Ferrara, Via Ariosto 35, 44122 Ferrara, Italy
\email{aneeqa.mehrab@unife.it} \and
Institute of Experimental Epileptology and Cognition Research, University of
Bonn Medical centre, Venusberg-Campus 1, 53127
Bonn, Germany
\email{petkojan@uni-bonn.de}
}
\maketitle              
\begin{abstract}
  We exploit the mathematical modeling of the visual cortex mechanism for
  border completion to define custom filters for CNNs. We see
  a consistent improvement in performance, particularly in accuracy, when our
  modified LeNet 5 is tested with occluded MNIST images.
  
\end{abstract}

\section{Introduction}
Visual perception has evolved as a fundamental tool for living organisms to
extract information from their surroundings and adapt their behavior. 
However, encoding visual information presents several challenges.
One major issue is occlusion, i.e. an object's outline is partially hidden by
an obstacle. In such cases, correctly identifying the object is far from
straightforward.
%
One functional solution %
is the orientation selectivity mechanism in the mammalian brain,
\cite{Hubel1969AnatomicalDO,Niell2008HighlySR}. This mechanism allows to
construct a percept of \textsl{orientation} at every point of a perceived
image, and it is implemented via a specific anatomical object, the
\textsl{hypercolumn} \cite{Hubel1974UniformityOM}. The hypercolumn is, at its
core, a group of cells, referred to as \textsl{columns}, each exhibiting
maximal response to a specific orientation realizing what is called
orientation \textsl{tuning}. This ultimately led to more sophisticated
mathematical modeling \cite{Hoffman1989TheVC}, later on developed by others
including \cite{Petitot1999VersUN,Bressloff2002TheVC} (see also
\cite{Citti2014NeuromathematicsOV} and refs. therein).
When  an image is processed, the hypercolumn's response is driven by the
column corresponding to the dominant orientation at that location, and this
information is relayed to higher cortical areas. Understanding how columnar
orientation selectivity emerges and how differently tuned columns interact has
been a major area of research since the discovery of cortical hypercolumns,
from experimental \cite{Pfeffer2013InhibitionOI}, \cite{Vinje2000SparseCA},
computational \cite{BernezTimn2022SynapticPC}, \cite{Goetz2021ActiveDE},
\cite{Kirchner2021EmergenceOL}, and theoretical perspectives
\cite{Tal1997TopologicalSI}, \cite{Schottdorf2015RandomWG}
\cite{Lindeberg2017NormativeTO}.
One crucial property enabled by the hypercolumnar structure is \textsl{contour
integration}. First observed and investigated by the Gestalt school of
psychology under the name of "law of good continuation"
\cite{Mather2006FoundationsOP}, this property allows the human (mammalian)
brain to reconstruct occluded contours building up from the information
present in the visible portion of the image \cite{Field1993ContourIB}. This
reconstruction can be understood in terms of neighbouring \textsl{association
fields}, which integrate aligned orientation cues scattered throughout the
input, and produce smooth, coherent contours \cite{field}. This effect is, in
fact, the behavioural counterpart of the hypercolumnar horizontal
connectivity, i.e. the interneuronal connections that excite (or inhibit)
collinear (or orthogonal) columns in adjacent hypercolumns. This
mechanism effectively resolves the occlusion problem, as partially hidden
contours can be \textsl{induced} from their visible portions, allowing the
complete object to be fully perceived, see the %
seminal work \cite{Marr1981DirectionalSA}, and later on in \cite{citti},
\cite{mumford},  \cite{Petitot1999VersUN},  \cite{Hoffman1989TheVC} for mathematical
modeling.

In this work we are interested in understanding if this biological mechanism
can be implemented computationally in a neural network model to effectively
increase its robustness against image occlusion during a classification task.
The parallels between the cortical connectivity and the neural network is
currently a lively field of research, \cite{Pemberton2024CerebellardrivenCD},
\cite{Poirazi2003PyramidalNA}, with bioinspired solutions leading to
significant improvements in a variety of tasks
\cite{Pagkalos2023LeveragingDP}, \cite{LeCun1998GradientbasedLA}. Building on
these and other previous results \cite{Petkovic2024SpontaneousEO} we show
how LeNet 5, a CNN known for its strong similarities with the mammalian
visual pathway, can benefit from the addition of biologically inspired
directional filters, closely resembling the receptive field observed for the
cortical hypercolumns. We show that incorporating these filters significantly
improves the model performance when classifying \textsl{occluded images}, i.e.
images where a black, oriented striping has been applied, despite the model
being trained on an unmodified dataset, thus realizing a contour integration,
very similar to the human one \cite{field}. Notably, this performance
improvement does not stem from a simple increase in parameter count but
appears to result from the specific geometric features of the introduced
filters.

The structure of the paper is as follows.
In Sec. \ref{sec-meth}, we recall briefly the mathematical modeling introduced
in \cite{Petkovic2024SpontaneousEO}, \cite{Fioresi2023ANP}, which is the base
for our subsequent treatment. We also describe the bioinspired deep learning
architecture, where we use predefined filters to mimic the action of the
vector fields on a contour and the border detection operator. We also present
our occluded dataset, obtained from MNIST, with diagonal occlusions, that we
shall use for our testing. It is important to remark that at no stage of our
training or validation, we employ such occluded dataset, reserving it only for
testing. In Sec. \ref{sec-res}, we discuss the results, namely we compare the
accuracies obtained by vanilla LeNet 5 with the ones achieved with our
enhanced CNN with border integration capabilities.

\section{Methodology}\label{sec-meth}
In this section we describe %
the mathematical modeling of border
perception as introduced in \cite{Fioresi2023ANP} and then %
the filter implementation mimicking the vector fields employed
for the border completion. At the end, %
we explain the
occluded dataset that we have created for our experiments.

\medskip
{\bf Mathematical Model of border perception.}
Following \cite{Fioresi2023ANP}, we think of an image as a smooth function 
$I \colon D \to \mathbb{R}$, where $D \subset \R^2$ corresponds to a receptive
field in the retina or equivalently in the visual cortex V1.
We now give a key definition, which is the base for our subsequent bioinspired
CNN.

\begin{definition}
\label{def::theta}
{\rm Let $I \colon D \to \mathbb{R}$ be a smooth function, and $\regu(D) \subseteq D$ 
the subset of the regular points of $I$, i.e. the points with non vanishing gradient.
We define the \textit{orientation map of $I$} as}
$$
    \Theta \colon \regu(D) \to S^1 \qquad
    (x,y) \mapsto \Theta(x,y) \, = \, 
\mathrm{argmax}_{\theta \in S^1} \{ Z(\theta)I(x,y)\}
$$
\end{definition}
where
$$
Z:=   Z(x,y,\theta) \, =\, -\sin\theta\, \partial_x + \cos\theta\, \partial_y,
$$
is called the \textit{orientation vector field}.

\begin{figure}[h!]
  \centering
  \includegraphics[width=0.9\textwidth]{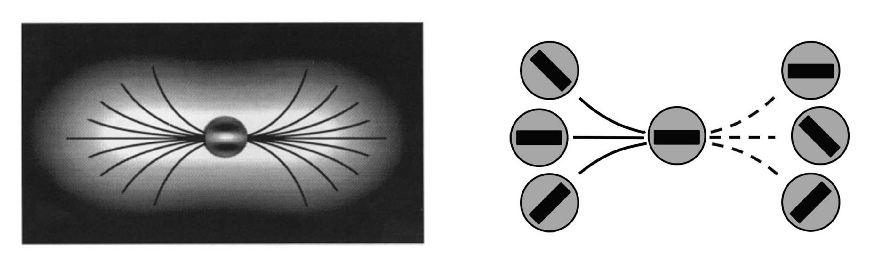}
  \caption{The association field.}
  \label{fig:field}
\end{figure}

Let $I:\dom \lra \R$ be as above and $(x_0,y_0)\in \dom$ a regular point. Then,
there there exists a unique $\theta_{x_0,y_0} \in S^1$ for which the function
$\zeta_{x,y}: S^1 \longrightarrow \R$, $\zeta_{x,y}(\theta):= Z \, I(x,y)$
attains its maximum. Hence, the map $\Theta$ 
is well defined and differentiable. This is an
immediate consequence of the implicit function theorem (see
\cite{Fioresi2023ANP} for more details). Notice that the locality of the
operator $Z$ (viewing vector fields as operators on functions) mirrors the
locality of the hypercolumnar anatomical connections.
In \cite{Fioresi2023ANP}, we show that, employing $Z$ to obtain a differential
constraint via a contact distribution, we can define a sub-Riemannian geodesic
problem. The geodesic curves solutions to this problem appear close to the ones
experimentally observed in \cite{field}, see Fig. \ref{fig:field}.

This highlights the key role of $Z$ and suggest the implementation of custom
filter replicating the orientation map $\Theta$.

\begin{figure}
\begin{center}
  \includegraphics[width=.6\textwidth]{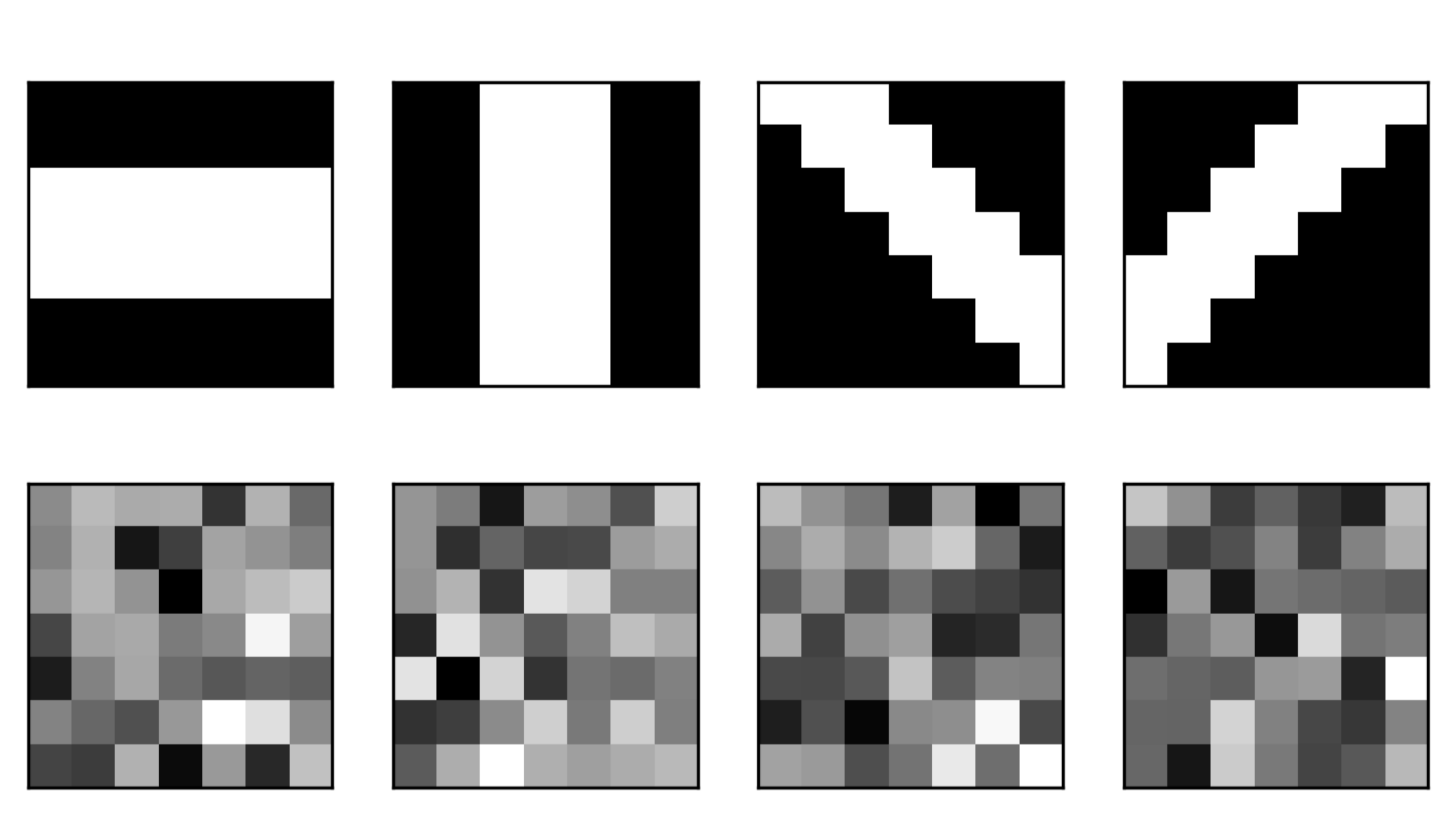}
\end{center}
  \caption{
    Oriented (top row) and random (bottom row) filters.} %
  \label{fig:filters}
\end{figure}

\medskip
{\bf Deep Learning Models.} 
We take into exam three CNN models: LeNet 5 (Vanilla LeNet) and two modified
LeNet 5, that we call BorderNet and RandomNet, in which four custom filters were added at the
beginning as a sequence of convolutions. For BorderNet these filters were created based on the
analogy with the visual cortex, each filter featuring one orientation -
horizontal, vertical, and both diagonals as in Fig
\ref{fig:filters} (top row). For RandomNet the filters are assigned randomly, as in Fig \ref{fig:filters}
(bottom row).
In both cases filters were defined with a
size of $7\times 7$ pixels and a stripe width of 3 pixels.
In each filter in BorderNet, the pixels belonging to the oriented stripe were set to the
  value of $1$, while the remaining background was set to $0$. In this fashion,
  a convolution with the filter would mimics the action of the vector field $Z$
as described above. 
RandomNet was obtained from Vanilla LeNet 
by adding four filters of the same dimension,
but with randomly generated pixel values. In this way, we create a CNN with the same
number of parameters as BorderNet, allowing us to check that the increased robustness
to occlusion is due to the added filters and not the number of parameters increase.

All the models were trained on the unoccluded MNIST training set, with ADAM
optimizer, learning rate of 0.001, 10 epochs, and a batch size of 64. The
testing was carried out using the occluded MNIST images, and model accuracy was
assessed for each combination of occlusion stripe width and spacing. 

\begin{figure}[h!]
  \centering
  \includegraphics[width=0.7\textwidth]{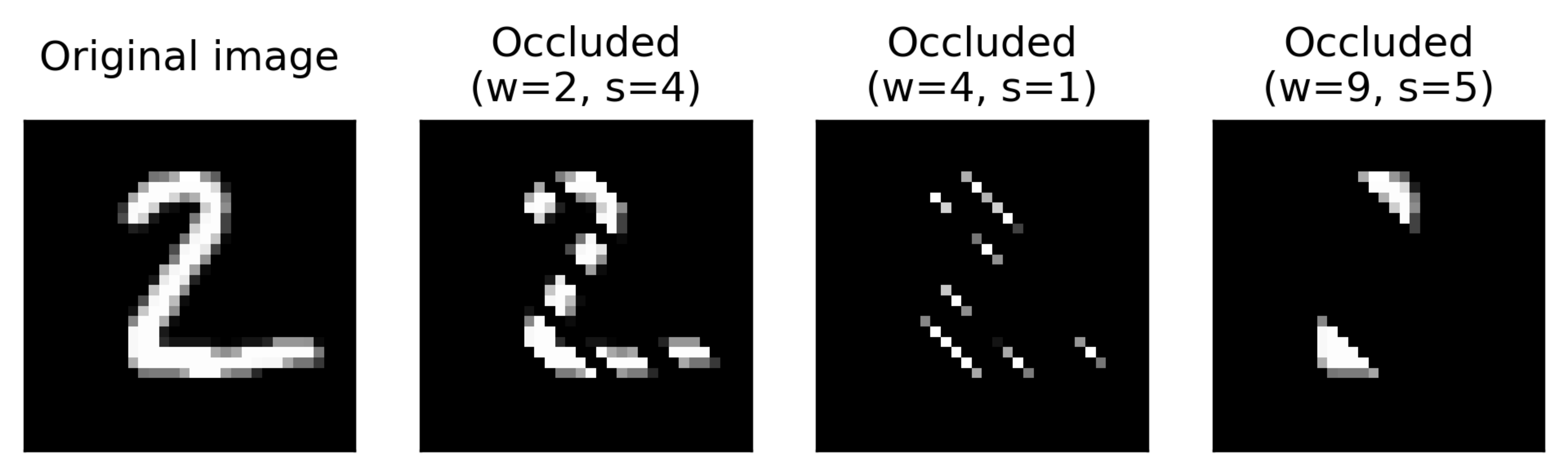}
  \caption{
A sample of occluded images used in the testing phase.}
  \label{fig:occlusions}
\end{figure}

{\bf Occlusions.}\label{occl-sec}
To test robustness of
our modified LeNet, i.e. BorderNet, we created a test dataset
consisting of occluded images as in Fig. \ref{fig:occlusions}.
As occlusion, we take a masking composed of diagonal straight stripes of
  width $w$ and inter-stripe spacing $s$. To conduct a thorough benchmarking, we
  test all the models on all the occlusions obtained from the pairwise 
  combination of $s, w \in [1,10]$. %

\section{Results}\label{sec-res}
In this section we report the results of testing of our three models
on occluded images, focusing on
the performance difference between BorderNet and the two control models (Vanilla LeNet
and RandomNet). %
It is important to remark once more that each CNN received the same training,
i.e. all CNNs were trained according to the method described in Sec. \ref{sec-meth} on
unoccluded MNIST images.

\begin{figure}[h!]
\centering
    \subfloat[\centering Vanilla LeNet]{{\includegraphics[scale=0.4]{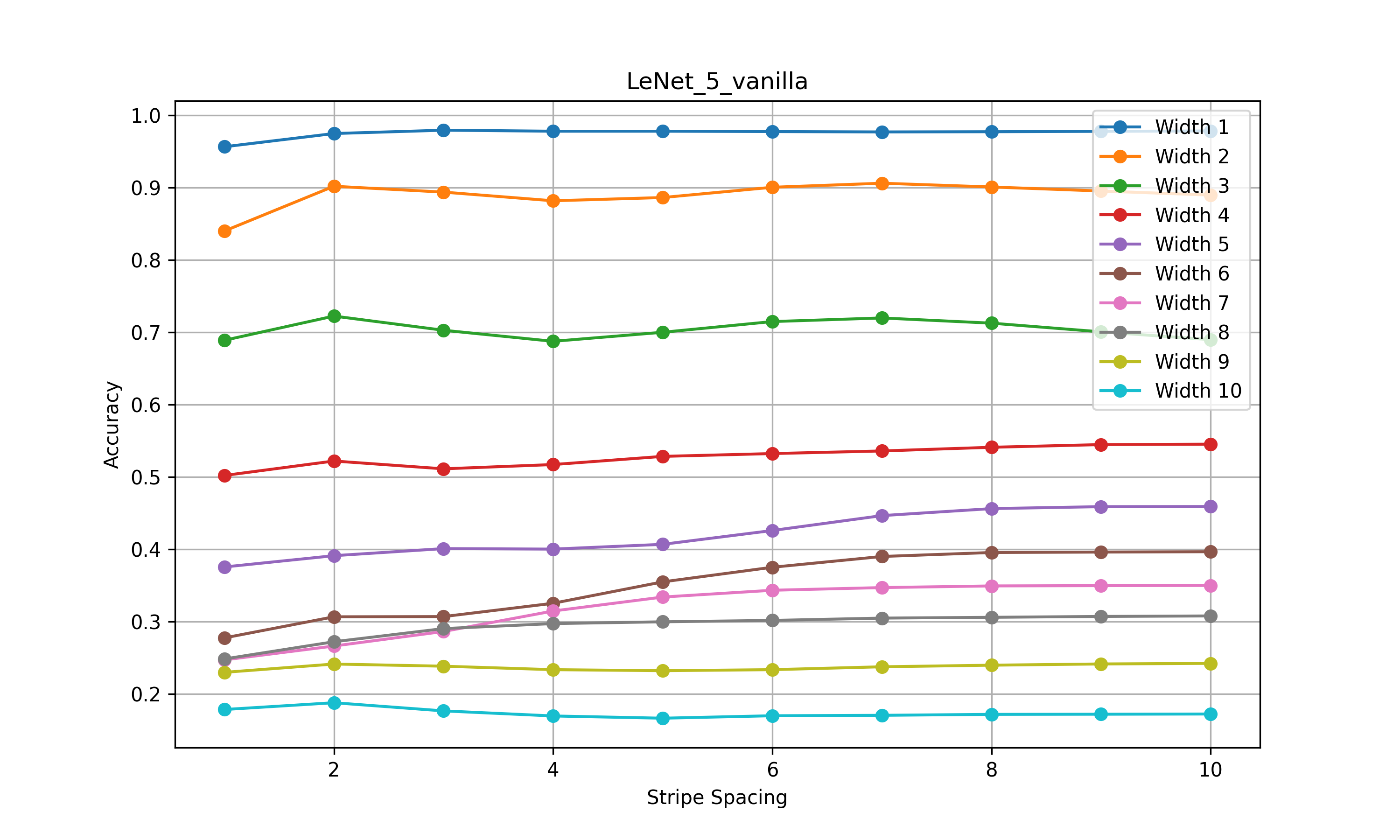} }}%
    \qquad
    \subfloat[\centering BorderNet]{{\includegraphics[scale=0.4]{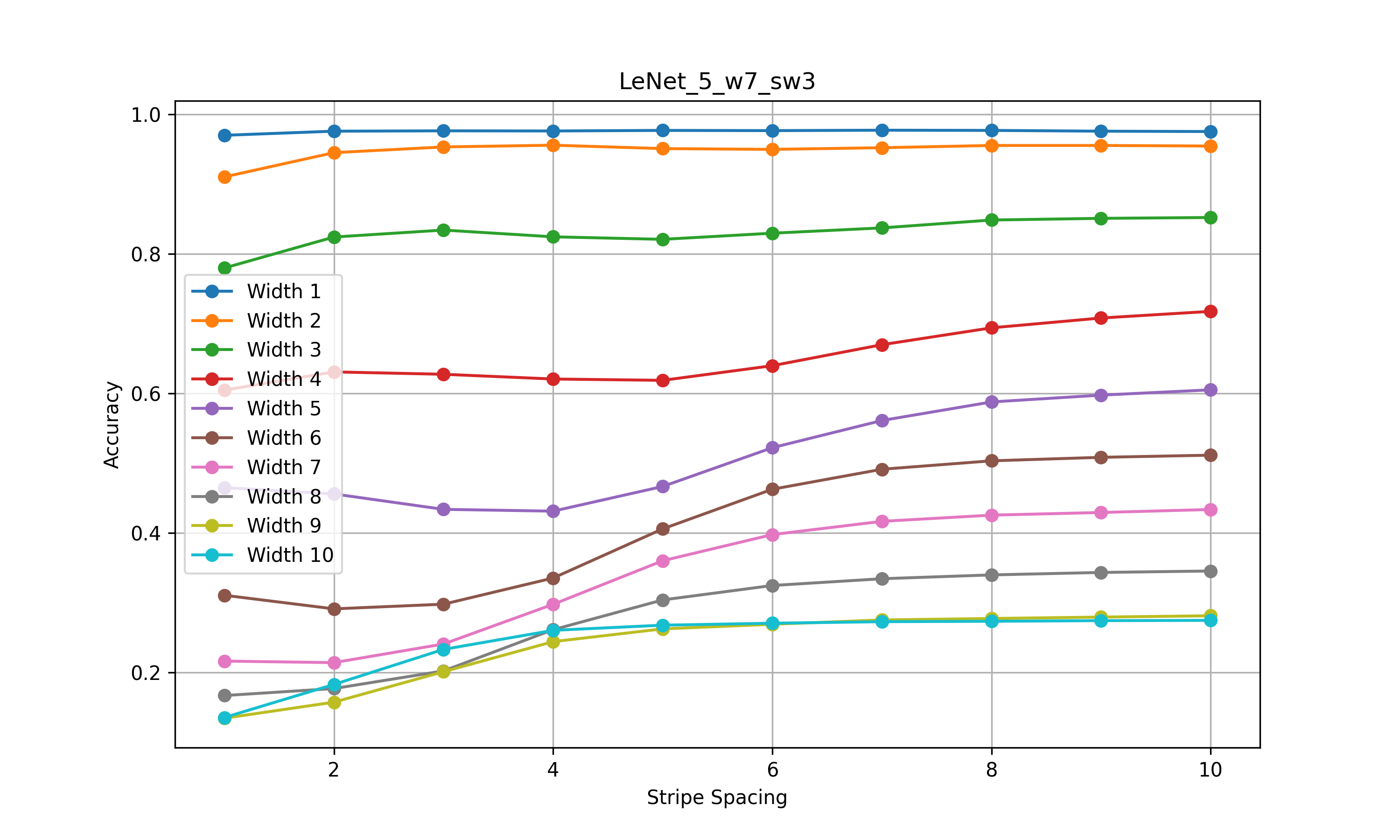} }}%
    \caption{Accuracies with occlusion stripe spacing and width for the
    Vanilla LeNet and BorderNet} 
    \label{fig:Lenet_perf}%
\end{figure}

\begin{figure}[h]
  \centering
  \includegraphics[scale=0.4]{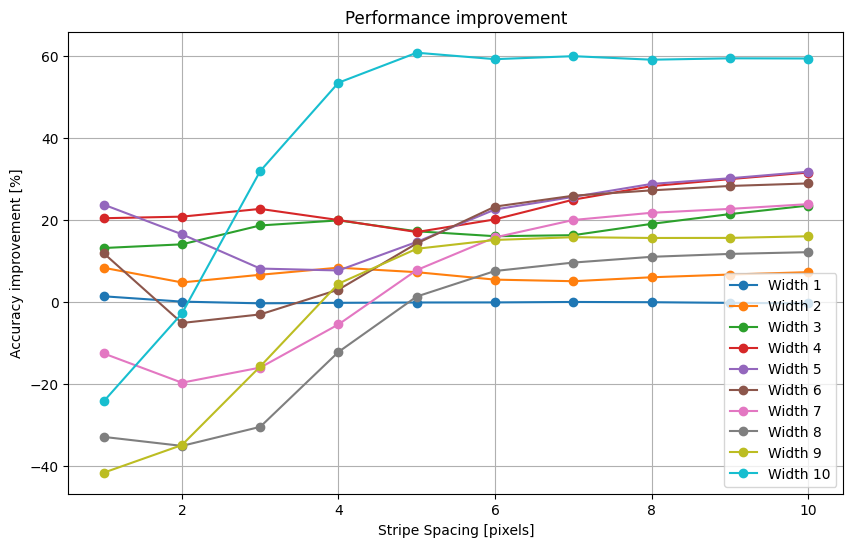}
\caption{Accuracy improvement of BorderNet with respect to Vanilla LeNet.}
  \label{fig:perf_comparison}
\end{figure}

\begin{figure}[!h]
  \centering
  \includegraphics[width=.7\textwidth]{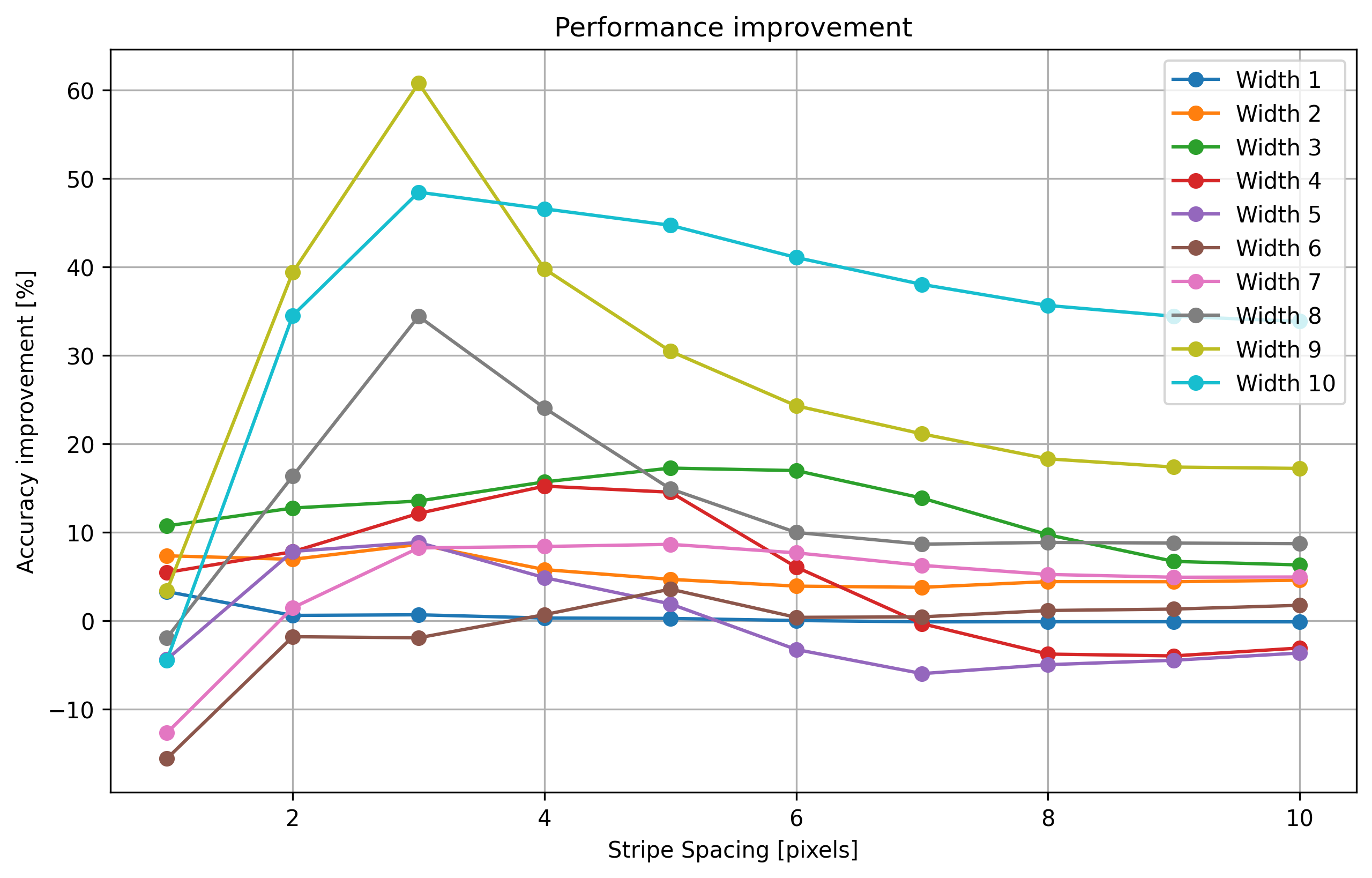}
\caption{Accuracy improvement of BorderNet with respect to RandomNet.}
%
  \label{fig:perf_random_comparison}
\end{figure}

As we can see in Fig. \ref{fig:perf_comparison}, we have a significant
performance enhancement by introducing our bioinspired custom filters, mimicking
the action of the orientation vector field $Z$, %
instrumental in our mathematical modeling of border completion. 
In Fig. \ref{fig:perf_random_comparison}
we see a consistent improvement  when comparing BorderNet with RandomNet i.e. Vanilla LeNet 
  equipped with random filters, %
  suggesting that higher accuracies cannot be attributed solely to an increase
in model parameter numbers. As a side remark, we notice that the negative
performances occur only for severe occlusions, similar to the two right
images in Fig. \ref{fig:occlusions}, thus not significant in the comparison.

\section{Conclusions}
Bioinspired custom filters improve the performance of CNNs, when tested on
dataset showing adverse ambient conditions (occlusions). Our proof of concept
shows the potential of a general improvement in performance of CNNs, by the use
of suitable custom filters, inspired by biology and by its mathematical
modeling. We expect to confirm our preliminary findings by testing and verifying
in the near future these results with more types of occlusions and different
types of benchmark datasets.

\begin{credits}
\subsubsection{\ackname} 
This research was funded by CaLISTA CA 21109, CaLIGOLA MSCA-2021-SE-01-101086123, MSCA-DN CaLiForNIA—101119552, PNRR MNESYS,the PNRR National Center for HPC, Big Data, and Quantum Computing, SimQuSec; INFN Sezione Bologna, Gast initiative and GNSAGA Indam.

\subsubsection{\discintname}
The authors have no competing interests.
\end{credits}


\bibliographystyle{splncs04}
\bibliography{biblio}

\end{document}